\documentclass[runningheads]{llncs}
\usepackage{latexsym}
\usepackage[ruled, lined, linesnumbered, commentsnumbered, longend]{algorithm2e}

\usepackage{amsmath}
\usepackage{amssymb}
\usepackage{setspace}
\usepackage{subfig}
\usepackage{comment}
\usepackage{booktabs}
\usepackage{graphicx}
\usepackage{wrapfig}
\usepackage{comment}

\DeclareMathOperator{\Tr}{Tr}

\newcommand{\argmax}[1]{\underset{#1}{\arg \max}\;}

\usepackage[T1]{fontenc}

\usepackage[utf8]{inputenc}

\usepackage{microtype}

\newcommand{\matr}[1]{\boldsymbol{\uppercase{#1}}}

\usepackage{graphicx}
\begin{document}
\title{Unsupervised Alignment of \\ Distributional Word Embeddings}
\titlerunning{Unsupervised Alignment}
%
\author{A{\"i}ssatou Diallo\inst{1}, Johannes F{\"u}rnkranz\inst{2}}
\authorrunning{A{\"i}ssatou Diallo \and Johannes F{\"u}rnkranz}
%
\institute{Department of Computer Science, University College London, United Kingdom \and Computational Data Analytics, FAW,
Johannes Kepler University Linz, Austria \\
\email{a.diallo@ucl.ac.uk, juffi@faw.jku.at}}
\maketitle              
\begin{abstract}
Cross-domain alignment plays a key role in tasks ranging from image-text retrieval to machine translation. The main objective is to associate related entities across different domains. Recently, purely unsupervised methods operating on monolingual embeddings have successfully been used to infer a bilingual lexicon without relying on supervision. However, current state-of-the art methods only focus on point vectors although distributional embeddings have proven to embed richer semantic information when representing words. This paper investigates a novel stochastic optimization approach for aligning word distributional embeddings. Our method builds upon techniques in optimal transport to resolve the cross-domain matching problem in a principled manner.
We evaluate our method on the problem of unsupervised word translation, by aligning word embeddings trained on monolingual data. We present empirical evidence to demonstrate the
validity of our approach to the bilingual lexicon induction task across several language pairs.

\end{abstract}

\section{Introduction}

Word embedding alignment is a fundamental Natural Language Processing task that aims at finding the correspondence between two sets of word embeddings. Word embeddings are vectorial representations of words capable of capturing the context of a word in a document, semantic and syntactic similarity as well as its relation to other words. Therefore, each embedding space exhibits different characteristics based on the semantic differences in the source of information provided as input. However, it has been first observed in \cite{mikolov2013exploiting} that continuous word embeddings exhibit similar structures across languages, even for distant ones such as English
and Vietnamese. For this reason, the task of aligning two clouds of points is a crucial problem in this specific setting.

Often, the set of embeddings to align are in different languages, i.e., we face the
task of cross-lingual alignment. Loosely speaking, given a source-target language pair, the goal is to find a mapping that goes from the embedding space of one language to the embedding space of the other language. An example related to Natural Language Processing (NLP) is the task of unsupervised word translation. In this setting, the learning process can be seen as a generalization of the unsupervised cross-domain adaptation problem \cite{sun2016return,mahadevan2018unified,ben2006analysis,gopalan2011domain}.

\begin{figure*}[ht]
\centering

\subfloat[Point vector embedding alignment. \label{fig:point_v}]{\includegraphics[scale=0.2]{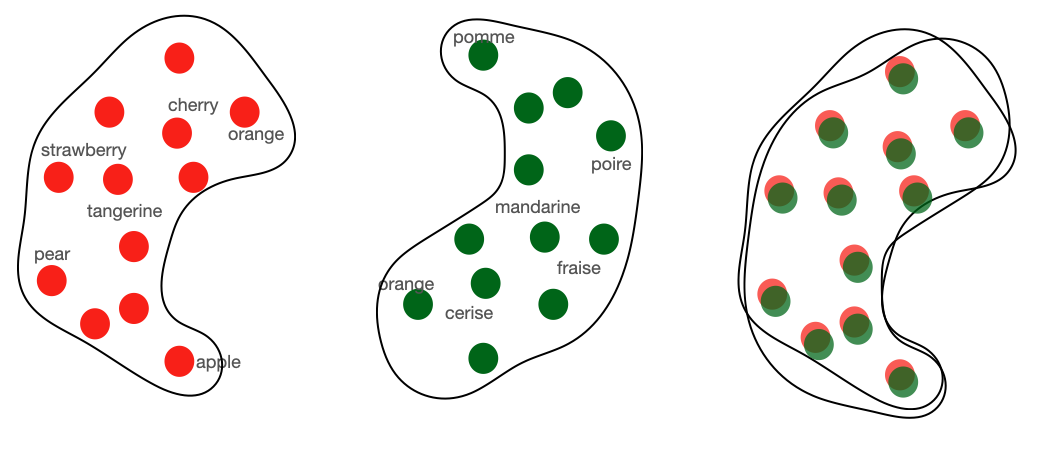}}
\hfill
\subfloat[Distributional vector embedding alignment. \label{fig:gauss_v}]{\includegraphics[scale=0.2]{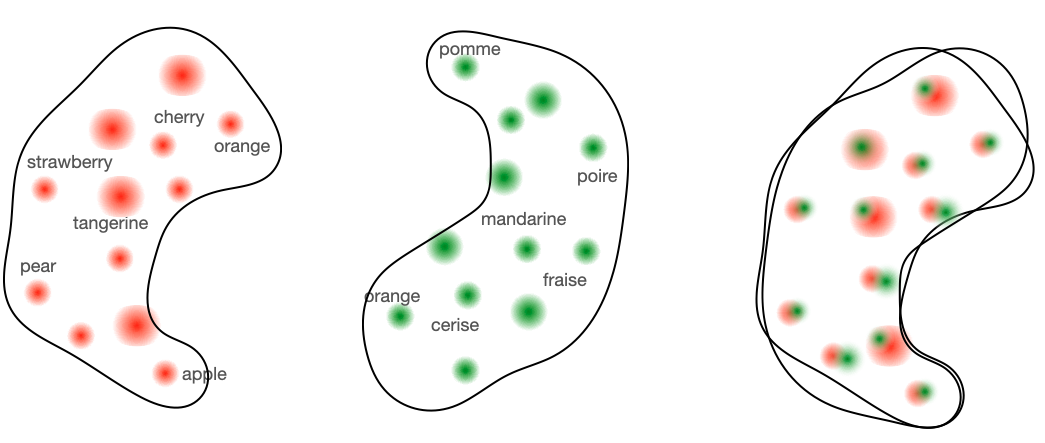}}

\caption{Unsupervised embedding alignment for two clouds of points in two different languages (English and French.)}
\end{figure*}

Several 
early studies have relied on supervision from a bilingual dictionary in the form of few anchor points in order to induce the learning of the mapping \cite{artetxe2018generalizing,joulin2018loss,jawanpuria2019learning}.
However, recently many unsupervised approaches have been proposed and have obtained compelling results \cite{zhang2017adversarial,grave2019unsupervised,alvarez2019towards,lample2018word,artetxe2018robust,zhang2017earth,jawanpuria2020geometry}. The unsupervised approaches frame the problem as a distance minimization between distributions using various distances, adversarial training, or domain adaptation. Generally speaking, all these methods build on the observation that mono-lingual word embeddings, or distributed representations of words, show similar geometric properties across languages. 
Another key point is the nature of the representation. 
Like other types of embeddings, word embeddings develop in two directions: point embeddings and probabilistic embeddings. 

\emph{Point embeddings} are powerful and compact representations that deterministically map each word into a single point in a semantic space, 
where the semantic similarity and other symmetric word relations are effectively captured by the relative position
of points. Despite these positive properties, this projection into a single point in the embedding space brings also several limitations. Most importantly, it has been shown that a single point vector struggles to naturally model entailment among words (e.g., animal entails dog but not vice versa) or other asymmetric relations. Moreover, point vectors are typically compared by dot products, cosine-distance or Euclidean distance, which are not well suited for 
carrying asymmetric comparisons between objects (as is necessary to represent relations such as inclusion or entailment).
Asymmetries can reveal hierarchical structures among words that can be  crucial in knowledge representation and reasoning \cite{roller2014inclusive}. Additionally, the point vector representation fails to express the uncertainty about the concepts associated with a specific word.

\begin{wrapfigure}{r}{0.5\textwidth}
  \begin{center}
    \includegraphics[width=0.48\textwidth]{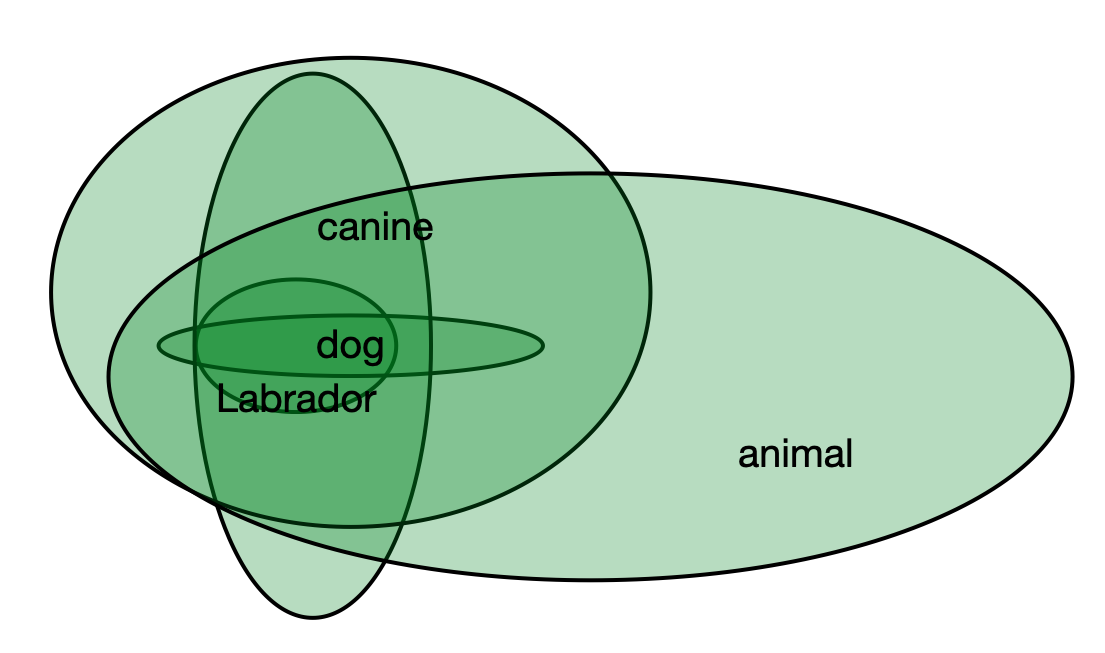}
  \end{center}
  \caption{Illustration of diagonal variances. Each word is defined by the position of its mean vector in the space and the dispersion is indicated by the variance. The more specific word \textit{Labrador} has a smaller variance than the more general categories \textit{animal} or \textit{canine}.}
\end{wrapfigure}
 
On the other hand, \emph{distributional embeddings} represent each word as a probability distribution, such as a Gaussian. Such a representation is innately more expressive having the ability to additionally capture semantic uncertainties of words (as their geometric shapes) to represent words more naturally and more accurately than point vectors \cite{vilnis2014word}. They allow mapping each word to soft regions in space in a manner that facilitates the modeling of uncertainty, inclusion and entailment.
Nevertheless, all the approaches for unsupervised alignment of word embedding focused on point vector. 

In this paper, we propose an approach for aligning embedding spaces for a source and a target language in an unsupervised manner that is suited for a large set of embeddings.  In particular, our algorithm shares similarities with the work of \cite{grave2019unsupervised} where a non-linear transformation and an alignment between two point clouds are jointly learned. Experiments show the validity of the proposed approach on the bilingual lexicon induction benchmark.

The paper is organized as follows: we first discuss related works that deal with point-vector and distributional embedding models as well as alignment of word embeddings with different degrees of supervision. Then, we formulate the problem and introduce the necessary notation used throughout the paper. Finally, we present our experimental setup and discuss the results obtained.

\section{Motivation and Related Work}

In this section, we briefly review the relevant state-of-the-art in this area, starting first with point-based and distributional embeddings in NLP, and moving then to the problem that we study in this paper, namely the alignment of word embeddings, where we briefly recapitulate supervised and unsupervised approaches. 

\subsection{Point-based word embeddings} One of the key problems in machine learning and natural language processing has been computing meaningful representation for high-dimensional complex data. This has been an active research area, from the traditional non-neural isometric embeddings \cite{brown1992class} \cite{blitzer2006domain}  to the more recent and complex methods \cite{Mikolov13distributedrepresentations}\cite{mikolov2013efficient} \cite{pennington2014glove}. And the most widely used algorithms for learning point-based word embeddings are the continuous bag of words and skip-gram models \cite{mikolov2013efficient}\cite{mikolov2013efficient}, which use a series of optimization methods such as negative sampling and hierarchical softmax \cite{mnih2008scalable}.
Another approach for learning word embeddings is through factorization of word co-occurrence matrices such as GloVe
embeddings \cite{pennington2014glove}. This
mechanism of matrix factorization has been proved to be intrinsically linked to skip-gram and negative sampling.

\subsection{Probabilistic Embedding} The work of \cite{vilnis2014word} established a new trend  in the representation learning field
by proposing to embed words as probability distributions in $\mathbb{R}^d$. In fact, recognizing that the point-based world struggles to naturally model entailment among words (e.g., animal entails cat but not the reverse) or other asymmetric relations, probabilistic embedding emerges as a method to capture uncertainties of words, which can better capture word semantics
and to express asymmetrical relationship more naturally (than dot product or
cosine similarity in the point-based approach). 
Representing objects in the latent space as probability distributions allows more flexibility in the representation and even express multi-modality. 
In fact, point-vector embeddings can be considered as an extreme case of probabilistic embeddings, namely a Dirac distribution, where the uncertainty is collapsed into a single point. 

\begin{figure*}[ht]
\centering
\includegraphics[scale=0.3]{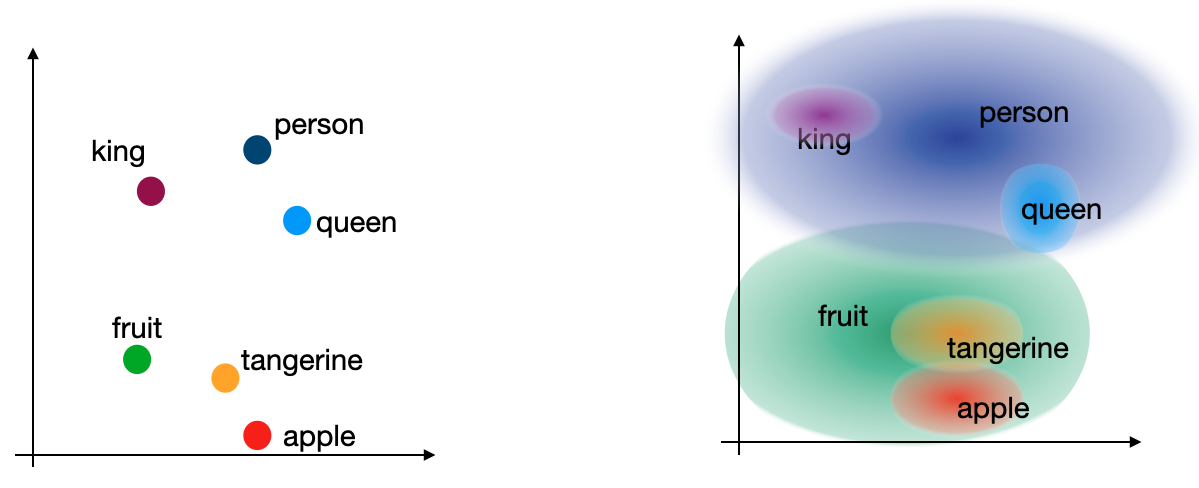}
\caption{Broader and more common terms have a wider dispersion than more specific ones. This characteristic is lacking in point-vector embeddings.}
\end{figure*}

\subsection{Minimally supervised alignment of word embeddings}
As stated earlier, word embeddings allow representing word relations in a metric space. Learning the projection of a word embedding space for a given language into another embedding space is useful in many applications, in particular in aligning vocabularies for different languages. 
Learning these cross-lingual mappings has initially been done using seed dictionaries. In fact, most early works
assumed some, albeit minimal, amount of parallel data \cite{mikolov2013exploiting,dinu2014improving,gaddy2016ten}. \cite{mikolov2013exploiting} proposes a mapping from one space to the other based on the least-squares objective whereas \cite{gaddy2016ten,smith2017offline,artetxe2018generalizing} aim at finding an orthogonal transformation. Other works fall under the minimally supervised category but aim at finding a common space on which to project both sets of embeddings \cite{faruqui2014improving,lu2015deep}.

\subsection{Fully unsupervised alignment of word embeddings}
In recent works in the area, it has been shown that fully-unsupervised
methods are able to perform on par
with their supervised counterparts. The first unsupervised bilingual alignment approaches \cite{miceli-barone-2016-towards} \cite{zhang2017adversarial}
\cite{lample2018word} 
were based on Generative Adversarial Networks (GANs) \cite{NIPS2014_5ca3e9b1}. These methods learn a linear transformation to minimize the divergence between a target
distribution (e.g. Spanish word embeddings) and a source distribution (the English word embeddings projected into the Spanish space).  In the recent literature, a range of unsupervised approaches that do not rely on the use of GANs has been
proposed \cite{artetxe2018generalizing} \cite{grave2019unsupervised}. Our approach relates more to those methods. \cite{artetxe2018generalizing} introduced a very simple, related initialization method that is, like our proposal, also based on
Gromov-Wasserstein distances between nearest neighbors: they use these second-order
statistics to build a seed dictionary directly by aligning nearest neighbors words across languages. \cite{alvarez2019towards} propose to
learn the doubly stochastic $\matr{Y}$, the matrix that determines the mapping between the words of the languages to be aligned, as a transport
mapping between the metric spaces of the words in the
source and the target languages. They optimize the
Gromov-Wasserstein (GW) distance, which measures how distances between pairs of words are
mapped across languages. In brief, \cite{alvarez2019towards} learn a linear transformation to minimize
Gromov-Wasserstein distances of distances between nearest neighbors, in the absence of cross-lingual supervision. 

Another line of work of interest 
attempts to solve the unsupervised alignment problem as a domain adaptation task \cite{sun2016return}. Their formulation searches the optimal permutation matrix for a limited number of items, specifically the 20000 most frequent  over the space of doubly stochastic matrices. They rely on a Riemannian solver that allows exploiting the geometry of the doubly stochastic manifold. Empirically,
the proposed algorithm outperforms the GW algorithm for learning bilingual mappings. Nevertheless, their approach is computationally more expensive.

However, all these approaches rely on point-vectors. In this paper, we argue that an unsupervised approach for aligning Gaussian embeddings can be beneficial because these types of embeddings have been proven to encode relations that normal point-vector fails to encode and could be particularly well suited for low-resource languages. 

\section{Approach}
In this section, we describe the unsupervised alignment problem and the solution strategy for dealing with Gaussian embeddings.

\subsection{Problem Formulation}
In the cross-lingual alignment problem, we are given a pair of source-target languages with vocabularies $V_x$, with $|V_x| = n$ and $V_y$, with $|V_y|= m$, respectively. These vocabularies are represented by word embeddings $\matr{X} \in \mathbb{R}^{n \times d}$ and $\matr{Y} \in \mathbb{R}^{m \times d}$. The goal of the problem in its classical form is to find a mapping between the set of source embeddings and target embeddings without parallel data. In this work, we tackle the problem of finding a mapping between sets of embeddings from a pair of languages but the inputs are not point-vectors. 
A Gaussian embedding can be seen as a generalization of point embeddings. Concretely, Gaussian embeddings are the result of representing data points as probability distributions, namely Gaussian measures in $\mathbb{R}^d$. Each Gaussian representation $w \sim \mathcal{N}(\mu, \Sigma)$ is a tuple of a mean $\mu \in \mathbb{R}^d$ (the location vector) and a covariance matrix $\Sigma \in \mathbb{S}^{d}$, the set of positive semi-definite $d \times d$ matrices. The covariance matrix can be seen as the dispersion that represents the uncertainty around the position of the location vector.
In this work, we focus on diagonal Gaussian embeddings, which are most used in the literature on probabilistic embeddings, but our approach can easily be extended to the general case with little effort.

The problem, then becomes, given a pair of sets of Gaussian embeddings from a source language $\mathcal{X}$ represented by $\matr{M}_x \in \mathbb{R}^{n \times d}$ and $\mathbf{\Sigma}_x \in \mathbb{R}_+^{n \times d}$ and from a target language $\mathcal{Y}$ $\matr{M}_y \in \mathbb{R}^{n \times d}$ and $\mathbf{\Sigma}_y \in \mathbb{R}_+^{n \times d}$, find a mapping $T: \mathcal{X} \xrightarrow{} \mathcal{Y}$ such that $T(x_i \in \matr{M}_x) \approx y_j \in \matr{M}_y$. 

In the next section, we begin by discussing the solution by \cite{grave2019unsupervised} and then present our adaptation to deal with Gaussian embeddings. 

\subsection{Orthogonal Procrustes}

The problem of finding a linear mapping between two clouds of matched vectors is known as \emph{Procrustes}. In the classical form, it is described as:
\begin{equation*}
    \min_{\matr{W \in \mathbb{R}^{d \times d}}}\|\matr{X}\matr{W} - \matr{Y}\|^2_F
\end{equation*}
where $\matr{W}$ is the learned mapping and $\|\cdot\|_F$ is the Frobenius norm.
This technique has been successfully applied in different fields, from analyzing sets of 2D
shapes to learning a linear mapping
between word vectors in two different languages with
the help of a bilingual lexicon \cite{mikolov2013exploiting}.
Constraints on the mapping $\matr{W}$ can be further imposed
to suit the geometry of the problem. 
An appropriate choice of the space for the mapping $T$ represented by $\matr{W}$ in the general case will be the space of the orthogonal matrices (rotations and reflections). Hence the problem becomes \emph{Orthogonal Procrustes}:
\begin{equation}
  \min_{\matr{Q \in \mathcal{O}_d}}  \|\matr{X}\matr{Q} - \matr{Y}\|^2_F
 \label{eq:procrustes}
\end{equation}
where $\mathcal{O}_d$ is the space of orthogonal matrices defined as :
\begin{equation}
    \mathcal{O}_d = \{ \matr{W} \in \mathbb{R}^{d \times d} | \matr{W}^T\matr{W=\matr{I}}\}. 
\end{equation}

The key advantage is that this problem has a closed-form solution. In fact, given the singular value decomposition of $\matr{X}\matr{Y}^T$ in $\matr{U}\matr{D}\matr{V}^T$, the optimal solution is 
\begin{equation}
\matr{Q}_{opt}=\matr{U}\matr{V}^T.
\end{equation}

\subsection{Wasserstein Procrustes}

However, the eq. \eqref{eq:procrustes} represents the supervised alignment problem, in which the learner is given a pair of sets of embedding correctly matched. If we generalize the problem, to the case in which the learner does not have access to a pair of matching embeddings, the problem at hand becomes:
\begin{equation}
    \min_{\matr{Q \in \mathcal{O}_d}, \matr{P} \in \mathcal{P}_d}  \|\matr{X}\matr{Q} - \matr{P}\matr{Y}\|^2_F
\label{eq:w_procrustes}
\end{equation}

In this general case, the permutation matrix $\matr{P}$ that represents the matching is also unknown. \cite{grave2019unsupervised} tackle this problem by jointly learning $\matr{P}$ and $\matr{W}$. While the overall problem is non-convex and computationally expensive, they propose an efficient stochastic algorithm to solve the problem and a convex relaxation which is used as an initialization for their algorithm. This convex relaxation, namely the Gold-Rangarajanng \cite{gold1996graduated} relaxation is a convex approximation of the NP-hard matching problem and can be solved with the Frank Wolfe algorithm. Loosely speaking, once an initial transformation is obtained, it is used for learning the singular value decomposition. 
Then, the authors propose a stochastic approach in which a batch of vectors is sampled from both languages, at each step $t$. This is motivated by the fact that the dimension of the permutation matrix $\matr{P}$ scales quadratically with the number of points $n$. The approach consists in alternating the full minimization of eq. \eqref{eq:w_procrustes} in $\matr{P}$ and a gradient-based update in $\matr{Q}$. 

\subsection{Wasserstein Procrustes for Gaussian Embedding}
In order to adapt the learning problem for Gaussian distributions as inputs, we re-frame the problem described by eq. \eqref{eq:w_procrustes}:
\begin{equation}
    \|\matr{M}_x\matr{R} - \matr{P}\matr{M}_y \|^2_F + \|\matr{\Sigma}_x - \matr{P}\matr{\Sigma}_y \|^2_F 
    \label{eq:ours}
\end{equation}
As stated earlier, $\matr{M}_x$ and $\matr{M}_y$ represent the location (mean) vectors of the source and target Gaussian embeddings respectively, whereas $\matr{\Sigma}_x$ and $\matr{\Sigma}_y$ are the diagonal covariance matrices. The transformation $\matr{R}$ is derived solely from the first term of the equation. The intuition comes from the fact that the covariances represent, from a geometrical point of view, the dispersion of the embeddings. However, the permutation matrix $\matr{P}$, which identifies the matching should be based also on the covariances. 
This is justified by the fact that mono-lingual embeddings exhibit similar geometric properties across languages and taking into account the covariances of the embedding acts as a regularization of the optimization problem.
Concretely, the permutation matrix $\matr{P}_t \in  \mathcal{P}_d$ at step $t$ is derived from $\matr{R}^T\matr{X}^T\matr{P}\matr{Y}$ and $\matr{\Sigma}_x^T\matr{\Sigma}_y$. The procedure is illustrated in Algorithm \ref{algo:unsup_alig}. 

\begin{algorithm}[t]

    \For{$t=1$ \KwTo $T$}{

    Draw $\matr{X}_t$ from $\matr{M_x}$ and $\matr{Y}_t$ from $\matr{M_y}$, of size $b$ \\
    Given the current $\matr{R}_t$, compute $\matr{P}_t$ between $\matr{X}_t$ and $\matr{Y}_t$ \\
    \quad $\matr{P}_t = \argmax{\matr{P} \in \mathcal{P}_b} \Tr(\matr{R}_t\matr{X}_t\matr{P}_t\matr{Y}_t) $ \\
    Compute the gradient $\matr{G}_t$ w.r.t $\matr{R}_t$: \\
    \quad $\matr{G}_t = -2\matr{X}_t \matr{P}_t \matr{Y}_t $ \\
    Gradient step: \\
    \quad $\matr{R}_{t+1} = (\matr{R}_t - \alpha \matr{G}_t) $ \\
    Project on the set of orthogonal matrices: \\
    \quad $\matr{R}_{t+1} = \prod_{\mathcal{O}_d}(\matr{R}_{t+1}) = \matr{U}\matr{V}^T $ \\
    
    \For{$i=1$ \KwTo $L$} { 
    
    Draw $\matr{X}_i$ from $\matr{X}_t\matr{R}_t$ and $\matr{Y}_i$ from $\matr{P}_t\matr{Y}_t$ \\
    Draw $\matr{C_x}_i$ from $\matr{\Sigma}_t$ and $\matr{C_y}_i$ from $\matr{P}_t\matr{\Sigma}_t$ \\
    \quad $\matr{P}_i = \argmax{\matr{P} \in \mathcal{P}_b}$
    $\Tr(\matr{C_x}_i^T\matr{C_y}_i) $ \\
    Compute the gradient $\matr{G}_i$ w.r.t $\matr{R}_i$: \\
    \quad $\matr{G}_i = -2\matr{X}_i \matr{P}_i \matr{Y}_i $ \\
    Gradient step: \\
    \quad $\matr{R}_{i+1} = (\matr{R}_i - \alpha \matr{G}_i) $ \\
    Project on the set of orthogonal matrices: \\
    \quad $\matr{R}_{i+1} = \prod_{\mathcal{O}_d}(\matr{R}_{i+1}) = \matr{U}\matr{V}^T $
    } }
    
    \caption{Unsupervised alignment of Gaussian embeddings}
    \label{algo:unsup_alig}
\end{algorithm}

The second term of eq. \eqref{eq:ours} is the contribution of the dispersion term to the Wasserstein distance of the Gaussian distributions. If we assume that the geometrical similarity of the embedding spaces is maintained across languages, then we can reasonably expect that corresponding embeddings in different languages will behave in the same way. As an example, we can consider words that describe a categorisation of elements such as the words "fruit" or "animal". We know that the Gaussian representations of these words have a greater dispersion than their more specific counterparts, such as "pear" or "dog". We can reasonably expect the same phenomenon to occur across languages. 
We propose to optimize this problem in steps:
\begin{itemize}
    \item First, learn an optimal orthogonal matrix $\matr{R}_t$ and permutation matrix $\matr{P}_t$ only using the means of the Gaussian embeddings. 
    \vspace{.25cm}
    \item Then, given this initial mapping and matching applications, refine the permutation matrix $\matr{P}_t$ with few iterations to match also the covariances and used this new learned $\matr{P}_i$ to derive $\matr{R}_i$.
\end{itemize}
The naive approach to optimize eq. \eqref{eq:ours} might be to add a term to take into account the covariances at step denoted by line 3 in Algorithm \ref{algo:unsup_alig}. However, the magnitude of the cost matrix derived from the covariances is too small, and we found that the best approach will be a nested gradient descent. First, we estimate optimal $\matr{P}$ and $\matr{R}$ only from the location vectors, then we refine them with a few iterations $L<<T$.

In order to quantitatively assess the
quality of our approach, we consider the problem of
bilingual lexicon induction for Gaussian embedding. In the next section, we describe the procedure to generate our mono-lingual probabilistic embeddings and we investigate the use of the covariance to learn the unsupervised alignment.

\section{Experiments}
In the following section, we present the experimental evaluation of our approach. Through this step, we seek to understand the impact of the covariance in the optimization dynamics and to evaluate the performance of our approach for the task of cross-lingual word embedding translation. 

\subsection{Data generation}
The first step for any unsupervised alignment algorithm is to provide the source and target embeddings. To the best of our knowledge, there aren't any trained Gaussian mono-lingual embeddings publicly available. The standard benchmark dataset for the cross-lingual is from \cite{lample2018word} trained with \emph{FastText} \cite{bojanowski2017enriching} on Wikipedia dumps and parallel dictionaries for 110 language pairs. The original Wikipedia dumps were not made available, which would have made it easier to retrain Gaussian embedding. 
We choose the following solution: we train a model using the method described in \cite{vilnis2014word} with the exception that the weights for the mean component of the model are initialized with \emph{FastText} embeddings. We fine-tune the embedding on Wikipedia dumps for each language for 3 epochs, with a learning rate $\lambda_r=0.05$ using Adagrad for optimization. We maintain the dimensionality of the \emph{FastText} embedding, i.e., $300$ dimensions. As generally done for language modeling, we keep only the tokens appearing more than 100 times in the text (for a total average number of
$210,000$ different words for all languages used). 

\subsection{Experimental setup}
After obtaining the required monolingual embeddings we proceed as follows: we first learn an alignment solely based on the means. This will be considered as the baseline that will allow us to appreciate the influence of the covariances in the computation of the alignment.
We follow the same training protocol as in \cite{grave2019unsupervised}. More precisely,
we perform 5 epochs and the batch size is doubled
at the beginning of each epoch while reducing the
number of iterations by a factor of 4. The first epoch of our method uses a batch
size of 500 and 5000 iterations. We  also use the
Sinkhorn solver of \cite{NIPS2013_cuturi} to compute approximate solutions of optimal transport problems, with a regularization parameter of 0.05. The number of iterations in the nested step is set at 2 and the learning rate is set at 0.1 times the learning rate used in the prior step.

Since the bilingual lexicon induction problem can be seen as a retrieval problem, the standard practice is to report the precision at one (P@1). 
As a criterion, we compute a direct nearest-neighbor search on the mean of the Gaussian embeddings. We tried computing a distance that will take into account the covariance matrix but we noticed that the impact on the P@1 score was negligible. 

Following \cite{alvarez2019towards}, we consider the top $n = 20,000$ most
frequent words in the vocabulary set for all the languages during the training stage. The inference is performed on the full vocabulary set.
The obtained results are summarized in Table~\ref{tab:result}. 

\begin{table*}[t]
  \centering
  \begin{tabular}{l|cc|cc|cc|cc}
\toprule
     & en-fr \hspace{.5cm} & fr-en & en-es  \hspace{.5cm} & es-en & en-de  \hspace{.5cm} & de-en & en-ru  \hspace{.5cm} & ru-en \\
    \midrule
    $\mu$ & 68.4  \hspace{.5cm} & 69.3 & 67.4  \hspace{.5cm} & 71.3 & 62.1  \hspace{.5cm} & 59.8 & \textbf{33.4}  \hspace{.5cm} & \textbf{49.7}  \\
    \textbf{$(\mu, \Sigma)$} & \textbf{70.5}  \hspace{.5cm} & \textbf{71.8} & \textbf{70.8} \hspace{.5cm} & \textbf{73.2} & \textbf{64.1}  \hspace{.5cm} & \textbf{60.1} & 29.6  \hspace{.5cm} & 41.2  \\
    \bottomrule
  \end{tabular}
  \vspace{.75cm}
  \caption{P@1 on five European languages: English, French, Spanish, German and Russian. Here "en-xx" refers to the average P@1 over multiple runs when English is the source language and xx is the target language. We notice, as expected that the performance is similar for closely related pairs of languages.}
  \label{tab:result}
\end{table*}

\subsection{Discussion}
In order to qualitatively assess the contribution of the covariance matrix, the results obtained considering the covariance matrices are compared to the ones without considering the covariance matrices. Overall, the performance improves when taking into account the covariances. This can also be explained by the fact that the terms containing the covariance act as a regularization. Due to the presence of a nested step, the computational time increases slightly compared to the point-vector case. However, the number of iterations in the nested loop is small, between 2 and 5, hence it is not a dramatic increase. 

One explanation for the improvement of the results when taking into account the covariances might be the refinement step. In fact, it has been observed in \cite{artetxe2018generalizing} and \cite{lample2018word} that refining the alignments improves the performance by a significant margin. 

A general observation is that similar pairs of languages have similar performance overall. However, some interesting points must be taken into account for the task of unsupervised bilingual dictionary induction:
\begin{itemize}
    \item \textbf{Impact of off-the-shelf embeddings}: We rely on the \textit{FastText} embedding for obtaining the embedding to align for different languages. However, \textit{FastText} is trained on approximately 16M sentences in Spanish and 1M sentences in English. The Gaussian embeddings are induced from the point vector and the performance can be explained by the quality of the starting embeddings. 
    \vspace{.25cm}
    \item \textbf{Impact of domain difference}: having a large monolingual corpus from similar domains across languages is of vital importance. In fact, it is known that when two corpora come from different topics or domains, the performance is extremely degraded. The domain dissimilarity computed by metrics such as the Jensen-Shannon divergence is significant. The term distribution is an important factor in Gaussian embeddings since the dispersion (the variance) is the direct result of the uncertainty inherent in the dataset. This is a factor that must be taken into account for aligning Gaussian embeddings in an unsupervised manner. 
    \vspace{.25cm}
     \item \textbf{Impact of language similarity}: The morphological typology of the language is a factor that should be considered. The main considered are: \textit{fusional}, \textit{agglutinative}, \textit{isolating}. A fusional language tends to form words by the fusion (rather than the agglutination) of morphemes so that the constituent elements of a word are not kept distinct. Notable examples are the Indo-European languages. An agglutinative language has words that are made up of a linear sequence of distinct morphemes and each component of meaning is represented by its own morpheme, for example, Finnish and Turkish. Finally, an isolating language is a natural language with no demonstrable genealogical relationship with other languages, examples are Vietnamese and Classical Chinese. The nature of the language pairs should be considered as it could have a bigger impact on unsupervised alignment for distributional embedding rather than the point-vector counterpart.
\end{itemize}

In general, our results are aligned with the performance shown by \cite{grave2019unsupervised} for the same retrieval criterion. It is worth noticing that the embeddings used are the result of quick fine-tuning, their quality is far lower than the \emph{FastText} embedding from the MUSE dataset \cite{lample2018word}. This is valid for all pairs of languages besides the coupling "en-ru". In this specific case, we observe that the performance of the covariance approach is worse than the alignment of the only means. This can be explained by the fact that English and Russian are distant languages and the relations expressed by the dispersion in the Gaussian embeddings in English might not correspond to the same relations in Russian. In fact, as stated earlier, the covariance matrix in Gaussian embeddings from a geometrical point of view corresponds to the uncertainty in the representation. Hence broad concepts have large variance and more focused concepts have a smaller variance or dispersion. 

Another explanation can be in the fact that the main assumption for unsupervised alignment approaches is that transformation is isomorphic. However, as shown in \cite{sogaard2018limitations} this is not true for all language pairs. And since our approach is based on enforcing similarity between concepts that might not share the same dispersion this might be an explanation for the poor performance in this specific language. A way to overcome this might be to provide a small seed dictionary and turn the problem into a minimally supervised one. Few key concepts that are geometrically related even in distant languages might work as landmark points. In the following section, we present the experimental evaluation of our approach. Through this step, we seek to understand the impact of the covariance in the optimization dynamics and to evaluate the performance of our approach for the task of cross-lingual word embedding translation.

\section{Conclusion}
This work presents a method to align Gaussian embeddings in high-dimensional space. Our approach is motivated by the fact that Gaussian embeddings have proven to possess characteristics that are not present in normal point-based vectors. We propose to include in the optimization of the Orthogonal Procrustes method via stochastic optimization a step that takes into account the difference between matched covariances. We show that our method performs better than the solely point-vector-based approach. However, we also observed that this approach might lead to a decrease in accuracy when the pair of languages considered is too distant. In fact, in that case, the approach might force distant concepts to have similar dispersion. 
In future work, we would like to extend this to deal with full covariance Gaussian embeddings as well as other elliptical embeddings and find a solution to overcome the issue of distant languages.

\section*{Acknowledgments}
This work has been supported by the German Research Foundation as part of the Research Training
Group Adaptive Preparation of Information from Heterogeneous Sources (AIPHES) under grant No. GRK 1994/1. The work was performed while Aïssatou Diallo was at Technische Universität Darmstadt.

\bibliographystyle{splncs04}
\bibliography{custom}

\end{document}